\documentclass[letterpaper, 10 pt, conference]{ieeeconf}

\usepackage{shellesc}

\IEEEoverridecommandlockouts %
\usepackage{amsmath,amsfonts,amssymb}
\usepackage{mathrsfs}
\DeclareMathOperator{\atan}{atan}
\usepackage{bbold} %
\usepackage{bbding} %
\newcommand{\medstar}{\text{\FiveStarOpen}} %

\usepackage{tabularx}
\usepackage{multirow}
\usepackage{siunitx}

\usepackage[export]{adjustbox}

\usepackage{graphicx}
\graphicspath{{figures/}}

\usepackage{caption}
\usepackage{xargs} %

\usepackage[style=ieee, backend=biber, bibencoding=utf8, maxbibnames=5, mincitenames=1, maxcitenames=2, url=false, doi=false, isbn=false, citestyle=numeric-comp, natbib=true, eprint=false]{biblatex}

\usepackage{bm} %
\usepackage{lipsum}

\usepackage{pgfplots}
\usepackage{pgfgantt}
\usepackage{pdflscape}
\pgfplotsset{compat=newest}
\pgfplotsset{plot coordinates/math parser=false}

\usepackage{tikz}
\usetikzlibrary{decorations.markings, positioning}
\usetikzlibrary{external}

\newlength\figureheight 
\newlength\figurewidth

\newcommand*\circleyes[1]{\tikzexternaldisable\tikz[baseline=(char.base)]{
		\node[shape=circle,draw,inner sep=1pt] (char) {#1};}\tikzexternalenable}
\newcommand*\circleno[1]{\tikzexternaldisable\tikz[baseline=(char.base)]{
		\node[shape=circle,draw,fill=gray!50,inner sep=1pt] (char) {#1};}\tikzexternalenable}

\newcommand*\circleyesnl{\circleyes{\footnotesize n}}
\newcommand*\circlenonl{\circleno{\footnotesize n}}

\newcommand\copyrighttext{%
	\scriptsize \textcolor{blue}{\textcopyright 2020 IEEE. Personal use of this material is permitted.  Permission from IEEE must be obtained for all other uses, in any current or future media, including reprinting/republishing this material for advertising or promotional purposes, creating new collective works, for resale or redistribution to servers or lists, or reuse of any copyrighted component of this work in other works}}
\newcommand\copyrightnotice{%
	\begin{tikzpicture}[remember picture,overlay]
	\node[anchor=north,yshift=-7.5pt] at (current page.north) {\fbox{\parbox{\dimexpr\textwidth-\fboxsep-\fboxrule\relax}{\copyrighttext}}};
	\end{tikzpicture}%
}

\title{\LARGE \bf Optimal Behavior Planning for Autonomous Driving: \\A Generic Mixed-Integer Formulation}
\author{Klemens Esterle$^{1*}$, Tobias Kessler$^{1*}$ and Alois Knoll$^{2}$%
	\thanks{$^{*}$These authors contributed equally to this work.}%
	\thanks{$^{1}$with fortiss GmbH, Research Institute of the Free State of Bavaria, Munich, Germany}%
	\thanks{$^{2}$with Chair of Robotics, Artificial Intelligence and Real-time Systems, Technical University of Munich, Munich, Germany}%
}

\begin{document}

\maketitle
\copyrightnotice
\thispagestyle{empty}
\pagestyle{empty}

\global\csname @topnum\endcsname 0
\global\csname @botnum\endcsname 0

\newcommand{\figurename}{Fig. }

\newcommand {\vect} {\boldsymbol}
\newcommand {\matr} {\boldsymbol}

\newcommand{\state} {\vect{x}}
\newcommand{\stateSpace} {\vect{\mathcal{X}}}
\newcommand{\beliefstate} {\vect{b}}
\newcommand{\contr} {\vect{u}}
\newcommand{\contrSpace} {\vect{\mathcal{U}}}
\newcommand{\meas} {\vect{y}}
\newcommand{\procNoise}{\vect{w}}

\newcommand {\cov}  {\matr{\Sigma}}

\newcommand{\stateNoDelta}{\hat\state}
\newcommand{\contrNoDelta}{\hat\contr}
\newcommand{\procNoiseNoDelta}{\hat\procNoise}

\newcommand{\abc}[2][\empty]{%
  \ifthenelse{\equal{#1}{\empty}}
    {no opt, mand.: \textbf{#2}}
    {opt: \textbf{#1}, mand.: \textbf{#2}}
}

\newcommand {\noiseu} {\procNoise}
\newcommand {\covu} {\matr{\Sigma_{\noiseu,}}}

\newcommand {\noiseuNoDelta} {\procNoiseNoDelta}
\newcommand {\covuNoDelta} {\matr{\Sigma_{\noiseuNoDelta,}}}

\newcommand {\defnoiseu}[1][\empty]{
 \ifthenelse{\equal{#1}{\empty}}
    {\noiseu\sim N(0,\covu)}
    {\noiseu_{#1}\sim N(0,\covu_{#1})}
}

\newcommand {\defnoiseuNoDelta}[1][\empty]{
 \ifthenelse{\equal{#1}{\empty}}
    {\noiseuNoDelta\sim N(0,\covuNoDelta)}
    {\noiseuNoDelta_{#1}\sim N(0,\covuNoDelta_{#1})}
}

\newcommand {\covm} {\matr{R}}
\newcommand {\noisem} {\vect{\nu}}
\newcommand {\defnoisem}[1][\empty]{
 \ifthenelse{\equal{#1}{\empty}}
    {\noisem\sim N(0,\covm)}
    {\noisem_{#1}\sim N(0,\covm_{#1})}
}

\newcommand{\stateB}{\vect{\xi}}
\newcommand{\contrB}{\vect{\nu}}
\newcommand{\procNoiseB}{\vect{\omega}}

\newcommand{\AB}{\mathcal{A}}
\newcommand{\BB}{\mathcal{B}}
\newcommand{\WB}{\mathcal{W}}
\newcommand{\costStateB}{\mathcal{Q}}
\newcommand{\costContrB}{\mathcal{R}}
\newcommand{\covStatesB}{\mathcal{S}_{\state}}
\newcommand{\covProcNoiseB}{\mathcal{S}_{\procNoise}}

\newcommand{\FB}{\mathcal{F}}

\newcommand{\cct}{\vect{t}}
\newcommand{\ccsval}{s}
\newcommand{\ccT}{\matr{T}}
\newcommand{\ccsvec}{\vect{s}}

\newcommand{\costState}{\matr{Q}}
\newcommand{\costContr}{\matr{R}}
\newcommand{\feedbackMatrix}{\matr{K}}
\newcommand{\cost}{J}

\newcommand{\stateConstraintMatrix}{\matr{C}}
\newcommand{\stateConstraintVector}{\vect{c}}

\newcommand{\stateConstraintFunc}{c}

\newcommand{\contrConstraintMatrix}{\matr{D}}
\newcommand{\contrConstraintVector}{\vect{d}}

\newcommand{\contrConstraintFunc}{d}

\newcommand{\stateRef}{\state^{*}}
\newcommand{\contrRef}{\contr^{*}}

\newcommand{\stateDelta}{\Delta\state}
\newcommand{\contrDelta}{\Delta\contr}

\newcommand {\Comment}[1]{\textcolor{blue}{#1}}

\newcommand {\partialder}[4][\bigg]{\frac{\partial #2}{\partial #3}#1|_{#4}}
\newcommand {\partialdernoarg}[3][\bigg]{\frac{\partial #2}{\partial #3}#1}

\newcommand{\nat}{\mathbb{N}}
\newcommand{\real}{\mathbb{R}}
\newcommand{\compl}{\mathbb{C}}

\newcommand{\norm}[1]{\left\| #1 \right\|}

\newcommand{\half}{\frac{1}{2}}

\newcommand{\parenth}[1]{ \left( #1 \right) }
\newcommand{\bracket}[1]{ \left[ #1 \right] }
\newcommand{\accolade}[1]{ \left\{ #1 \right\} }
\newcommand{\pardevS}[2]{ \delta_{#1} f(#2) }
\newcommand{\pardevF}[2]{ \frac{\partial #1}{\partial #2} }

\newcommand{\vecii}[2]{\begin{pmatrix} #1 \\ #2 \end{pmatrix}}
\newcommand{\veciii}[3]{\begin{pmatrix}  #1 \\ #2 \\ #3	\end{pmatrix} }
\newcommand{\veciv}[4]{\begin{pmatrix}  #1 \\ #2 \\ #3 \\ #4	\end{pmatrix}}

\newcommand{\matii}[4]{\left[ \begin{array}[h]{cc} #1 & #2 \\ #3 & #4 \end{array} \right]}
\newcommand{\matiii}[9]{\left[ \begin{array}[h]{ccc} #1 & #2 & #3 \\ #4 & #5 & #6 \\ #7 & #8 & #9	\end{array} \right]}

\newcommand{\transp}{^{\intercal}}
\newcommand{\Reg}{$^{\textregistered}$}
\newcommand{\reg}{$^{\textregistered}$ }
\newcommand{\Tm}{\texttrademark}
\newcommand{\tm}{\texttrademark~}
\newcommand {\bsl} {$\backslash$}

\newtheorem{theorem}{Theorem}[section]
\newtheorem{lemma}[theorem]{Lemma}
\newtheorem{corollary}[theorem]{Corollary}
\newtheorem{remark}[theorem]{Remark}
\newtheorem{definition}[theorem]{Definition}
\newtheorem{equat}[theorem]{Equation}
\newtheorem{example}[theorem]{Example}
\newcommand{\insertfigure}[4]{ %
	\begin{figure}[htbp]
		\begin{center}
			\includegraphics[width=#4\textwidth]{#1}
		\end{center}
		\vspace{-0.4cm}
		\caption{#2}
		\label{#3}
	\end{figure}
}

\newcommand{\refFigure}[1]{\figurename \ref{#1}}
\newcommand{\refChapter}[1]{Chapter \ref{#1}}
\newcommand{\refSection}[1]{Section \ref{#1}}
\newcommand{\refParagraph}[1]{Paragraph \ref{#1}}
\newcommand{\refEquation}[1]{Eq. (\ref{#1})}
\newcommand{\refTable}[1]{Table \ref{#1}}
\newcommand{\refAlgorithm}[1]{Algorithm \ref{#1}}

\newcommand{\rigidTransform}[2]
{
	${}^{#2}\!\mathbf{H}_{#1}$
}

\newcommand{\code}[1]
 {\texttt{#1}}

\newcommand{\comment}[1]{\marginpar{\raggedright \noindent \footnotesize {\sl #1} }}

\newcommand{\clearemptydoublepage}{%
  \ifthenelse{\boolean{@twoside}}{\newpage{\pagestyle{empty}\cleardoublepage}}%
  {\clearpage}}

\newcommand{\etAl}{\emph{et al.}\mbox{ }}

\makeatletter
\def\ignorecitefornumbering#1{%
	\begingroup
	\@fileswfalse
	#1%
	\endgroup
}
\makeatother

\newcommand{\pxy}{p_\medstar}
\newcommand{\vxy}{v_\medstar}
\newcommand{\axy}{a_\medstar}
\newcommand{\uxy}{u_\medstar}
\newcommand{\xy}{\square_\medstar}
\newcommand{\deltat}{\Delta t}
\newcommand{\x}{p_x}
\newcommand{\y}{p_y}
\newcommand{\vx}{v_x}
\newcommand{\vy}{v_y}
\newcommand{\ax}{a_x}
\newcommand{\ay}{a_y}
\newcommand{\jx}{j_x}
\newcommand{\jy}{j_y}
\newcommand{\ux}{u_x}
\newcommand{\uy}{u_y}
\newcommand{\xfrontlb}{\underline{f_x}}
\newcommand{\yfrontlb}{\underline{f_y}}
\newcommand{\xfrontub}{\overline{f_x}}
\newcommand{\yfrontub}{\overline{f_y}}
\newcommand{\xyfrontlb}{\underline{f_\medstar}}
\newcommand{\xyfrontub}{\overline{f_\medstar}}
\newcommand{\squarefrontlb}{\underline{f_{\medstar}}}
\newcommand{\squarefrontub}{\overline{f_{\medstar}}}
\newcommand{\xfront}{f_{x}}
\newcommand{\yfront}{f_{y}}
\newcommand{\xyfront}{f_{\medstar}}
\newcommand{\xref}{p_{x,ref}}
\newcommand{\yref}{p_{y,ref}}
\newcommand{\vxref}{v_{x,ref}}
\newcommand{\vyref}{v_{y,ref}}
\newcommand{\orientation}{\theta}
\newcommand{\orientationlb}{\underline{\orientation}}
\newcommand{\orientationub}{\overline{\orientation}}
\newcommand{\region}{\rho} %
\newcommand{\possibleregion}{\varrho^r}
\newcommand{\umaxx}{\overline{\ux^r}}
\newcommand{\umaxy}{\overline{\uy^r}}
\newcommand{\uminx}{\underline{\ux^r}}
\newcommand{\uminy}{\underline{\uy^r}}
\newcommand{\uminxy}{\underline{\uxy^r}}
\newcommand{\umaxxy}{\overline{\uxy^r}}
\newcommand{\amaxx}{\overline{\ax^r}}
\newcommand{\amaxy}{\overline{\ay^r}}
\newcommand{\aminx}{\underline{\ax^r}}
\newcommand{\aminy}{\underline{\ay^r}}
\newcommand{\aminxy}{\underline{\axy^r}}
\newcommand{\amaxxy}{\overline{\axy^r}}
\newcommand{\noregionchange}{\Psi}
\newcommand{\noregionchangexpos}{\noregionchange_x^{+}}
\newcommand{\noregionchangexneg}{\noregionchange_x^{-}}
\newcommand{\noregionchangeypos}{\noregionchange_y^{+}}
\newcommand{\noregionchangeyneg}{\noregionchange_y^{-}}
\newcommand{\vregchange}{V}%
\newcommand{\vmin}{\underline{v}}
\newcommand{\vmax}{\overline{v}}
\newcommand{\amin}{\underline{a}}
\newcommand{\amax}{\overline{a}}
\newcommand{\umin}{\underline{u}}
\newcommand{\umax}{\overline{u}}
\newcommand{\fractionparametersone}{\alpha}
\newcommand{\fractionparameterstwo}{\beta}
\newcommand{\fractionparametersthree}{\gamma}
\newcommand{\fractionparametersfour}{\delta}
\newcommand{\polysinub}{\overline{\mathscr{P}_{\sin}^r}}
\newcommand{\polycosub}{\overline{\mathscr{P}_{\cos}^r}}
\newcommand{\polysinlb}{\underline{\mathscr{P}_{\sin}^r}}
\newcommand{\polycoslb}{\underline{\mathscr{P}_{\cos}^r}}
\newcommand{\polycosubone}{\overline{\chi_C}}
\newcommand{\polycosubtwo}{\overline{\psi_C}}
\newcommand{\polycosubthree}{\overline{\omega_C}}
\newcommand{\polysinubone}{\overline{\chi_S}}
\newcommand{\polysinubtwo}{\overline{\psi_S}}
\newcommand{\polysinubthree}{\overline{\omega_S}}
\newcommand{\polycoslbone}{\underline{\chi_C}}
\newcommand{\polycoslbtwo}{\underline{\psi_C}}
\newcommand{\polycoslbthree}{\underline{\omega_C}}
\newcommand{\polysinlbone}{\underline{\chi_S}}
\newcommand{\polysinlbtwo}{\underline{\psi_S}}
\newcommand{\polysinlbthree}{\underline{\omega_S}}
\newcommand{\polykappaub}{\overline{\mathscr{P}_{\kappa}^r}}
\newcommand{\polykappalb}{\underline{\mathscr{P}_{\kappa}^r}}
\newcommand{\polykappa}{\mathscr{P}_{\kappa}^r}
\newcommand{\polykappaminone}{\underline{\chi_\kappa}}
\newcommand{\polykappamintwo}{\underline{\psi_\kappa}}
\newcommand{\polykappaminthree}{\underline{\omega_\kappa}}
\newcommand{\polykappamaxone}{\overline{\chi_\kappa}}
\newcommand{\polykappamaxtwo}{\overline{\psi_\kappa}}
\newcommand{\polykappamaxthree}{\overline{\omega_\kappa}}
\newcommand{\notwithinenv}{e}
\newcommand{\envpolys}{\Lambda}
\newcommand{\envpolyline}{\lambda}
\newcommand{\envpolylinesegment}{l}
\newcommand{\envpolylinesegmentpointone}{a}
\newcommand{\envpolylinesegmentpointtwo}{b}
\newcommand{\deltacc}{o_p}
\newcommand{\deltaccall}{o_\medstar}
\newcommand{\deltaccfrontlblb}{o_{\underline{f}\underline{f}}}
\newcommand{\deltaccfrontlbub}{o_{\underline{f}\overline{f}}}
\newcommand{\deltaccfrontublb}{o_{\overline{f}\underline{f}}}
\newcommand{\deltaccfrontubub}{o_{\overline{f}\overline{f}}}
\newcommand{\obstacle}{\mathscr{o}}
\newcommand{\obstacleset}{\mathscr{O}}
\newcommand{\setofregions}{\mathscr{R}}
\newcommand{\timeintervall}{\mathscr{K}}
\renewcommand{\t}{k}
\newcommand{\StateMatrix}{\bm{A}}
\newcommand{\InputMatrix}{\bm{B}}

\begin{abstract}
Mixed-Integer Quadratic Programming (MIQP) has been identified as a suitable approach for finding an optimal solution to the behavior planning problem with low runtimes. Logical constraints and continuous equations are optimized alongside.
However, it has only been formulated for a straight road, omitting common situations such as taking turns at intersections. This has prevented the model from being used in reality so far.
Based on a triple integrator model formulation, we compute the orientation of the vehicle and model it in a disjunctive manner. That allows us to formulate linear constraints to account for the non-holonomy and collision avoidance. These constraints are approximations, for which we introduce the theory.
We show the applicability in two benchmark scenarios and prove the feasibility by solving the same models using nonlinear optimization.
This new model will allow researchers to leverage the benefits of MIQP, such as logical constraints, or global optimality.
\end{abstract}

\IEEEpeerreviewmaketitle

\section{Introduction}
\label{sec:introduction}

The scope of a behavior planning component for autonomous driving is calculating a plan to safely navigate through traffic reaching a desired state. 
This plan is usually defined as a sequence of future states or waypoints, which is then passed to a trajectory tracking controller. 
If those points cannot be followed, because they do not account for the non-holonomy of the vehicle, safety is at risk and eventually collision can occur.

Optimization-based methods incorporate a model of the kinematics, which is propagated for a given planning time horizon, and usually formulate constraints to account for feasibility and safety while constructing a cost function to account for comfort and other desired aspects.
Although local continuous optimization such as sequential quadratic programming (SQP) has been applied in reality \cite{Ziegler2014}, mixed-integer programming (MIP) offers multiple benefits that are favorable for optimal behavior planning. First, MIP will yield an optimal solution, whereas SQP only guarantees yielding a local optimum. Secondly, it allows logical and integer constraints to be incorporated, whereas they usually lead to numerical issues with continuous solvers. 
MIP cannot use a non-linear vehicle model, such as the bicycle model, as the resulting optimization problem is hard to solve. Past studies used second- or third-order integrators to mitigate that \cite{Frese2011, Qian2016}.

\begin{figure}[tb]
	\footnotesize
	\centering
	\def\svgwidth{0.6\columnwidth}
\begingroup%
  \makeatletter%
  \providecommand\color[2][]{%
    \errmessage{(Inkscape) Color is used for the text in Inkscape, but the package 'color.sty' is not loaded}%
    \renewcommand\color[2][]{}%
  }%
  \providecommand\transparent[1]{%
    \errmessage{(Inkscape) Transparency is used (non-zero) for the text in Inkscape, but the package 'transparent.sty' is not loaded}%
    \renewcommand\transparent[1]{}%
  }%
  \providecommand\rotatebox[2]{#2}%
  \newcommand*\fsize{\dimexpr\f@size pt\relax}%
  \newcommand*\lineheight[1]{\fontsize{\fsize}{#1\fsize}\selectfont}%
  \ifx\svgwidth\undefined%
    \setlength{\unitlength}{1526.89528097bp}%
    \ifx\svgscale\undefined%
      \relax%
    \else%
      \setlength{\unitlength}{\unitlength * \real{\svgscale}}%
    \fi%
  \else%
    \setlength{\unitlength}{\svgwidth}%
  \fi%
  \global\let\svgwidth\undefined%
  \global\let\svgscale\undefined%
  \makeatother%
  \begin{picture}(1,0.69219419)%
    \lineheight{1}%
    \setlength\tabcolsep{0pt}%
    \put(0,0){\includegraphics[width=\unitlength,page=1]{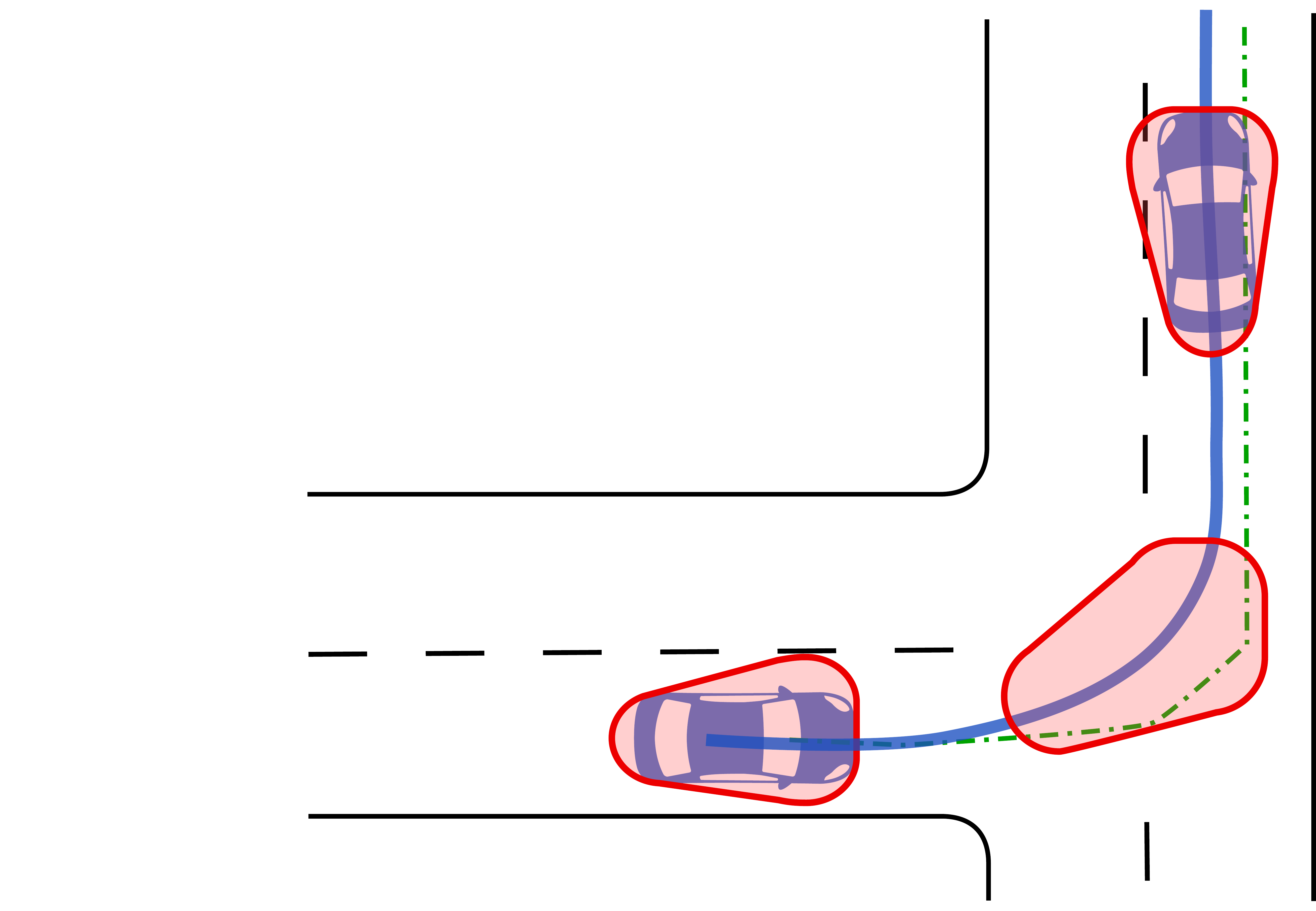}}%
    \put(0.09032713,0.5967112){\color[rgb]{0,0,0}\makebox(0,0)[lt]{\lineheight{1.25}\smash{\begin{tabular}[t]{l}Collision Shape Ego\end{tabular}}}}%
    \put(0.08907296,0.53745145){\color[rgb]{0,0,0}\makebox(0,0)[lt]{\lineheight{1.25}\smash{\begin{tabular}[t]{l}Reference Trajectory\end{tabular}}}}%
    \put(0,0){\includegraphics[width=\unitlength,page=2]{intro_T_intersection.pdf}}%
    \put(0.09032713,0.47778186){\color[rgb]{0,0,0}\makebox(0,0)[lt]{\lineheight{1.25}\smash{\begin{tabular}[t]{l}Optimized Trajectory\end{tabular}}}}%
    \put(0,0){\includegraphics[width=\unitlength,page=3]{intro_T_intersection.pdf}}%
    \put(0.09167439,0.41883882){\color[rgb]{0,0,0}\makebox(0,0)[lt]{\lineheight{1.25}\smash{\begin{tabular}[t]{l}Collision Shape Obstacle\end{tabular}}}}%
    \put(0,0){\includegraphics[width=\unitlength,page=4]{intro_T_intersection.pdf}}%
  \end{picture}%
\endgroup%

	\caption{Overview of our approach. From an arbitrary environment shape and a non-trackable reference trajectory, we compute an optimal motion. By over-approximating the collision shape of the vehicle with adapted constraints for different orientations, our model is globally valid and avoids arbitrary obstacles.}
	\label{fig:intro}
\end{figure}

However, the proposed methods can only be used on a small subset, namely straight roads, and become invalid in any other environment (roundabouts, intersections), or even during obstacle avoidance at low speeds, as the valid scope of the model formulation is limited. 
This poses the question of how to formulate a generic globally valid linearized model with correct vehicle dynamics in a MIP framework as sketched in \refFigure{fig:intro}.

We contribute a model applicable to the full-fledged range of orientations for the vehicle, featuring 
\begin{itemize}
	\item linear over-approximating collision constraints,
	\item linear non-holonomy constraints of the vehicle,
	\item the methodology to compute all model parameters by linear least-square fits and
	\item a detailed study of the feasibility of the model.
\end{itemize}

\section{Related Work}
\label{sec:related_work}

\begin{table*}[tb]
	\centering
	\vspace{0.15cm}
	\caption{Comparison of model approaches. We express linear functionals using \protect\circleyes{1}, quadratic using \protect\circleyes{2}, and all other non-linear functionals using \protect\circleyes{\normalsize n}. The counterparts \protect\circleno{1}, \protect\circleno{2}, and \protect\circleno{\normalsize n} express functionals that are not (and cannot be) implemented. We compare cost functionals that we assume to be desirable, mostly following \cite{Ziegler2014}. $j_\square$ denote the cost terms, $\kappa$ the curvature, $v$ the velocity, and $a$ the acceleration.}
	\scriptsize
	\begin{tabular}{l|l|l|l|l|l}
Source & \ignorecitefornumbering{\citet{Ziegler2014}} & \ignorecitefornumbering{\citet{Gutjahr2016a}} & \ignorecitefornumbering{\citet{Nilsson2016}} & \ignorecitefornumbering{\citet{Qian2016}} & \ignorecitefornumbering{\citet{Frese2011}} \\
\hline 
Problem formulation & SQP & QP for long, lat each & QP for long, lat each & MIQP & MILP \\
Model & triple integrator & Frenet bicycle model & double integrator & double integrator & double integrator \\
Reference frame & Cartesian, fixed & Frenet, streetwise & Cartesian, fixed & Cartesian, fixed & Cartesian, fixed \\
\hline 
Non-Holonomy Constraint & $\kappa$, \circleyesnl & $\kappa$, \circleyes{1} & $v_x, v_y$ are coupled, \circleyes{1} & $v_x, v_y$ are coupled, \circleyes{1} & $a_{lat}$, \circleyes{1} \\
Acceleration Constraint $|a|$ & \circleyes{2} & \circleyes{1} & \circlenonl & \circlenonl & approximated, \circleyes{1} \\
Collision Shape & disks & disks & road-aligned rectangle & road-aligned rectangle & road-aligned rectangle \\
Collision Check to & everything & everything & road-aligned rectangle & road-aligned convex polygon & non-convex road polygon \\
Collision Constraint & \circleyesnl & \circleyes{1} & \circleyes{1} & \circleyes{1} & \circleyes{1} \\
\hline 
$j_{\text{veloctiy}}$ & \circleyesnl & \circleyes{2} & \circlenonl & \circlenonl & \circlenonl \\
$j_{\text{acceleration}}$ & \circleyes{2} & \circleyes{2} & \circleyes{2} & \circleyes{2} & \circleyes{1}, \circleno{2} \\
$j_{\text{jerk}}$ & \circleyes{2} & \circleyes{2} & \circleno{2} & \circleyes{2}, see \ignorecitefornumbering{\cite{Burger2018}} & \circleno{2} \\
$j_{\text{yawrate}}$ & \circleyesnl & \circleyes{2} using $\dot{\kappa}$ \footnote{to curvature of reference line} & \circlenonl & \circlenonl & \circlenonl \\
$j_{\text{reference}}$ & \circleyesnl & \circleyes{2} & \circlenonl & \circlenonl & \circlenonl \\
\hline 
Validity & any road / orientation& any road / orientation & \multicolumn{3}{c}{straight road, aligned with road (orientation-wise)} \\
Multi-Agent & no& no& no& yes, see \ignorecitefornumbering{\cite{Burger2018}} & yes \\
\end{tabular}%
	\label{tab:model_comparism}
\end{table*}

\citet{Ziegler2014} proposed an optimal control problem, where they apply the cost functionals and constraints shown in \refTable{tab:model_comparism}.
They use a triple integrator as vehicle model, while bounding the curvature to account for non-holonomy. The Bertha-Benz drive showed the applicability of the triple integrator model if correctly constrained. Their approach yields a non-linear optimization problem, which is solved using SQP but only finds a local optimum. Motivated by that, approaches for maneuver planning \cite{Bender2015, Altche2018a} have focused on finding the correct maneuver in a preliminary step. However, these approaches usually rely on a geometric partitioning of the workspace and thus scale poorly, and cannot be extended to account for any interactive or cooperative planning.

\citet{Nilsson2016} introduce two quadratic problems (QP) for longitudinal and lateral control based on a linear double-integrator model. To account for non-holonomy, the authors use a linear inequality constraint that couples lateral and longitudinal velocity.
However, this is only valid for small yaw angles and will yield non-drivable trajectories at intersections and roundabouts for example, since the road curvature is not taken into account.

\citet{Gutjahr2016a} introduce two longitudinal and lateral QP in the Frenet frame based on the bicycle model. 
The model yields good results for driving in static environments. 
Similar to \citet{Ziegler2014}, this approach relies on a maneuver selection, as shown by \citet{Esterle2018}.
However, the transformation of all obstacles 
to local coordinates is costly and with an increasing number of obstacles outweighs the benefits of the fast QP.

\citet{Miller2018} base their work on \cite{Gutjahr2016a} and formulate two consecutive longitudinal and lateral programs using mixed-integer quadratic programming (MIQP) similar to \cite{Nilsson2016}, but in local street-wise coordinates instead. 
However, the approach cannot account for any model-based prediction or multi-agent planning. 
Also separating longitudinal and lateral control yields sub-optimal solutions and limits the solution space.

\citet{Frese2011} propose a double integrator based Mixed Integer Linear Programming formulation, leveraging its capability of yielding an optimal solution.
That comes with the fact that both only linear constraints and objective functions can be formulated. 
We thus analyzed the formulations in \refTable{tab:model_comparism} in terms of linear functionals \protect\circleyes{1}, quadratic \protect\circleyes{2}, and non-linear functionals \protect\circleyes{\normalsize n}. 
The non-holonomy is assured by bounding an approximation of the lateral acceleration.
However, that is only valid for small yaw angles, and yields a lot of invalid solutions \cite{Frese2012}. 
The collision checks are modeled with an arbitrary road polygon using a disjunctive collision check with convex polygons. They also propose to check for collision using rectangle-based vehicle-approximation of consecutive states, with each variant introducing a lot of invalid trajectories \cite{Frese2012}.

\citet{Qian2016} apply the double integrator to MIQP. Similar to \citet{Nilsson2016}, they model the non-holonomy by bounding the lateral velocity, which is only valid for straight roads and if the vehicle is road-oriented.
Even when avoiding an obstacle, the yaw angle may exceed 20 degrees, which can yield bends in the optimized trajectories that cannot be executed with a real vehicle. 
The work is thus limited to straight roads and straight driving, whereas turning at intersections is not possible.
With the quadratic cost function of a MIQP, differences to longitudinal or lateral values (such as acceleration) are only possible if the reference signal is zero (such as for acceleration). 
Thus, deviations from the absolute velocity cannot be expressed.
\citet{Burger2018} extend the work of \citet{Qian2016} to the cooperative setting by extending the state space to multiple agents. However, also the limitations of the formulation are inherited.

\citet{Kessler2019} formulate an MIP to select an optimal cooperative behavior for a set of agents. 
Instead of relying on a model within the optimization, they use motion primitives, which makes it applicable in arbitrary environments. 
However, with a narrow discretization, the optimization problem becomes large.

To summarize, no method currently exists to apply the advantages of MIP (global solution, logical constraints) to generic environments.

\section{Region-based Linearization Approach}
\label{sec:approximations}
Following \citet{Qian2016}, we model the vehicle as a third-order point-mass system with positions $\x(t)$, $\y(t)$, velocities $\vy(t)$, $\vy(t)$, and accelerations $\ax(t)$, $\ay(t)$ as states.
Jerk in both directions $\jx(t)$ and $\jy(t)$ are inputs of the model.
As we aim to formulate the vehicle model as linear constraints, all non-linearities have to be eliminated. 
In the following, we will introduce how we guarantee the validity of the model for all orientations and perform collision checks.

\begin{table}[b]
	\caption{parameters from fitting or preprocessing used throughout this paper. Region  dependency is denoted by $r$.}
	\centering
	\begin{tabularx}{8cm}{r|X} %
		Parameter & Description \\ 
		\hline 
		$\fractionparametersone^r$ & $x$ value of region $r$ lower region borderline\\
		$\fractionparameterstwo^r$ & $y$ value of region $r$ lower region borderline\\
		$\fractionparametersthree^r$ & $x$ value of region $r$ upper region borderline\\
		$\fractionparametersfour^r$ & $y$ value of region $r$ upper region borderline\\
		$\polycosub$ & polynome linear in $\vx$, $\vy$ upper-bounding $\cos(\orientation)$\\
		$\polycoslb$ & polynome linear in $\vx$, $\vy$ lower-bounding $\cos(\orientation)$\\
		$\polysinub$ & polynome linear in $\vx$, $\vy$ upper-bounding $\sin(\orientation)$\\
		$\polysinlb$ & polynome linear in $\vx$, $\vy$ lower-bounding $\sin(\orientation)$\\
		$\polykappalb$ & polynome linear in $\vx$, $\vy$ lower-bounding the $\kappa$\\
		$\polykappaub$ & polynome linear in $\vx$, $\vy$ upper-bounding the $\kappa$\\
		$\uminx$, $\umaxx$ & lower and upper jerk limit in direction $x$\\
		$\uminy$, $\umaxy$ & lower and upper jerk limit in direction $y$\\
		$\aminx$, $\amaxx$ & lower and upper acceleration limit in direction $x$\\
		$\aminy$, $\amaxy$ & lower and upper acceleration limit in direction $y$\\
		$\possibleregion$ & the region $r$ is allowed for the current scenario\\ 	
	\end{tabularx} 	
	\label{tab:nomenclature_fittingparameters}
\end{table}

\subsection{Discretized and Disjunctive Modeling of the Orientation}
\label{subsec:disjunctive_modelling}

\begin{figure}[t]
	\vspace{0.15cm}
	\footnotesize
	\centering
	\def\svgwidth{0.65\columnwidth}
	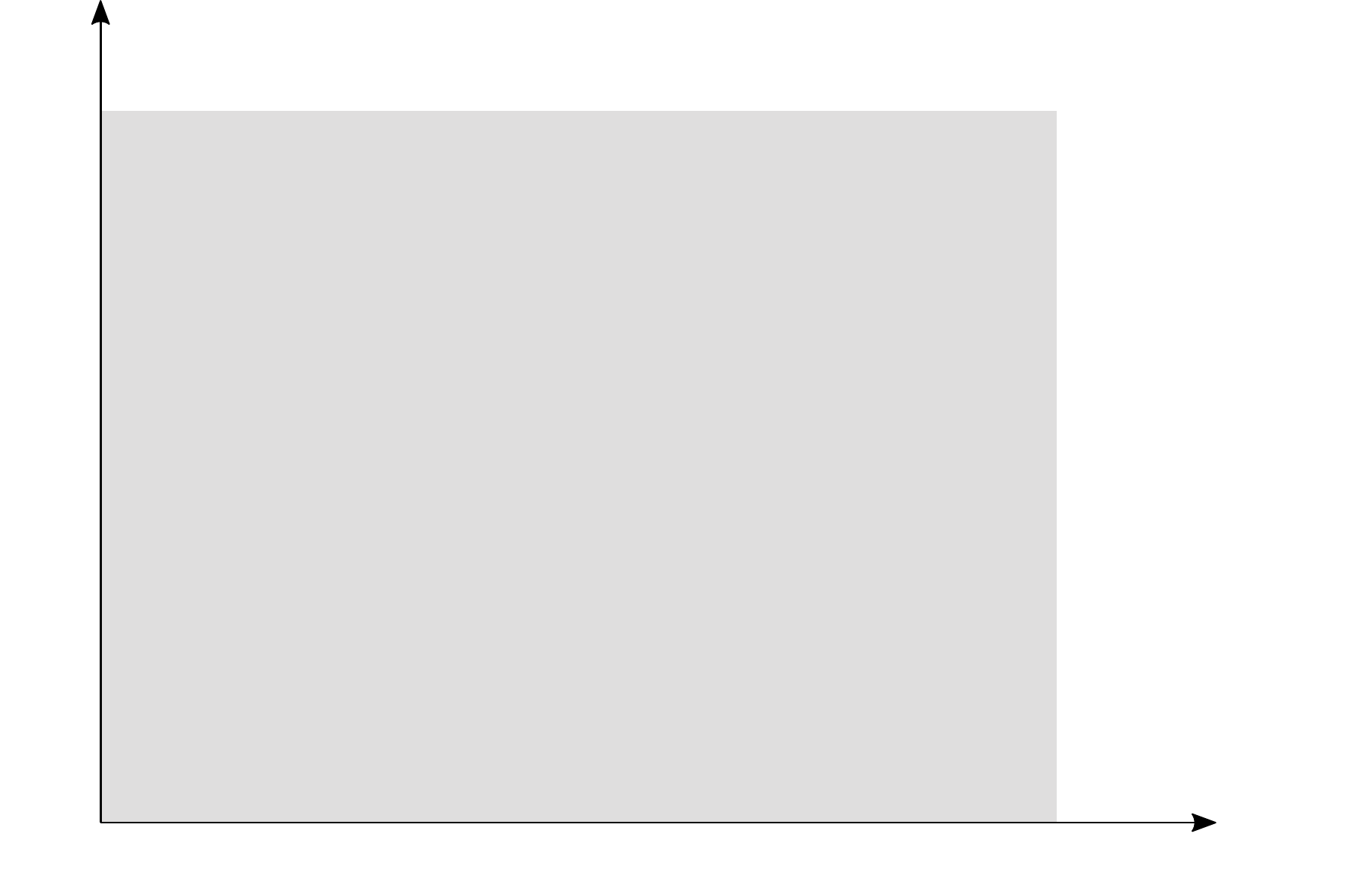
	\caption{Construction of region $i$ described through two lines $(0,0) - (\fractionparametersone, \fractionparameterstwo)$ and $(0,0) - (\fractionparametersthree, \fractionparametersfour)$ following \refEquation{eq:region_linear_equations}}
	\label{fig:region_construction}
	\vspace{-0.15cm}
\end{figure}

Although the vehicle's orientation $\orientation$ is not part of the state space, we will need it for a sufficient collision check in Cartesian coordinates within the optimization problem. 
We assume perfect traction and therefore neglect vehicle and tire slip.
It can be calculated using $\theta = \atan(v_y/v_x)$.
However, this equation is non-linear, as are the trigonometric operations 
\begin{subequations}
\begin{align}
	\sin(\theta) &= \sin(\atan(v_y/v_x)) \label{eq:trigonometrics_sin} \\
	\cos(\theta) &= \cos(\atan(v_y/v_x)) \label{eq:trigonometrics_cos}
\end{align}
\label{eq:trigonometrics_sin_cos}
\end{subequations}
are necessary to calculate the front axle position. To formulate collision constraints, we thus need to linearize \refEquation{eq:trigonometrics_sin_cos}. 
For our model to be valid for orientations $\orientation \in [0, 2\pi]$, we discretize the orientation by introducing \textit{regions} in the $(\vx, \vy)$ plane, see \refFigure{fig:region_construction}. 
The regions are defined by the area between two line segments starting at the origin.
Consequently, for every $(\vx, \vy)$ point within a region $i$, the following inequalities hold with region-dependent line parameters:
\begin{subequations}
\begin{align}
\fractionparametersone \vy &\geq \fractionparameterstwo \vx \label{eq:fractionlineone} \\
\fractionparametersthree \vy &\leq \fractionparametersfour \vx \label{eq:fractionlinetwo}
\end{align}
\label{eq:region_linear_equations}
\end{subequations}
Having done this, we will subsequently formulate model equations that are valid in each region with different parameters.
These are listed in \refTable{tab:nomenclature_fittingparameters}.

For driving comfort reasons, most motion planners limit the maximum longitudinal and lateral acceleration, deceleration, and jerk. 
From these desired values in the driving direction of the vehicle, we compute region-specific bounds in global $x$, $y$ coordinates.
This is done by rotating the original longitudinal and lateral limits along with the vehicle orientation. 
As the orientation angle, we chose the mean angle of the respective region. 
This ensures that we comply with the original bounds in driving direction in terms of absolute values and directions.

\subsection{Over-Approximating the Collision Shape}
\label{subsec:overapproximation_collision_shape}
\begin{figure}[t]
	\vspace{0.15cm}
	\footnotesize
	\centering
	\def\svgwidth{0.65\columnwidth}
\begingroup%
  \makeatletter%
  \providecommand\color[2][]{%
    \errmessage{(Inkscape) Color is used for the text in Inkscape, but the package 'color.sty' is not loaded}%
    \renewcommand\color[2][]{}%
  }%
  \providecommand\transparent[1]{%
    \errmessage{(Inkscape) Transparency is used (non-zero) for the text in Inkscape, but the package 'transparent.sty' is not loaded}%
    \renewcommand\transparent[1]{}%
  }%
  \providecommand\rotatebox[2]{#2}%
  \newcommand*\fsize{\dimexpr\f@size pt\relax}%
  \newcommand*\lineheight[1]{\fontsize{\fsize}{#1\fsize}\selectfont}%
  \ifx\svgwidth\undefined%
    \setlength{\unitlength}{232.20540925bp}%
    \ifx\svgscale\undefined%
      \relax%
    \else%
      \setlength{\unitlength}{\unitlength * \real{\svgscale}}%
    \fi%
  \else%
    \setlength{\unitlength}{\svgwidth}%
  \fi%
  \global\let\svgwidth\undefined%
  \global\let\svgscale\undefined%
  \makeatother%
  \begin{picture}(1,0.40051386)%
    \lineheight{1}%
    \setlength\tabcolsep{0pt}%
    \put(0,0){\includegraphics[width=\unitlength,page=1]{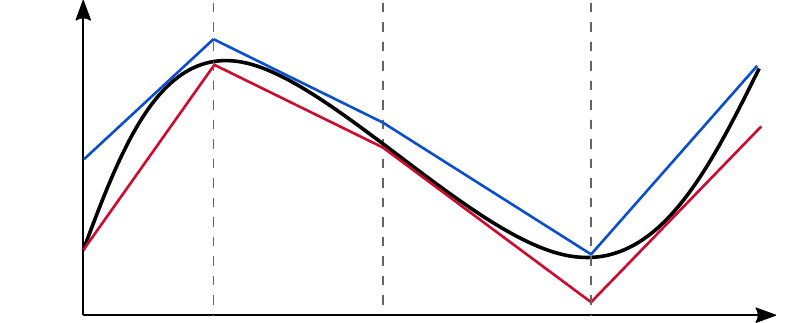}}%
    \put(-0.12143364,0.05334277){\color[rgb]{0.39215686,0.39215686,0.39215686}\makebox(0,0)[lt]{\lineheight{1.25}\smash{\begin{tabular}[t]{l}Region $1$\end{tabular}}}}%
    \put(0.27723353,0.35686741){\color[rgb]{0.39215686,0.39215686,0.39215686}\makebox(0,0)[lt]{\lineheight{1.25}\smash{\begin{tabular}[t]{l}Region $2$\end{tabular}}}}%
    \put(0.49286612,0.02521642){\color[rgb]{0.39215686,0.39215686,0.39215686}\makebox(0,0)[lt]{\lineheight{1.25}\smash{\begin{tabular}[t]{l}Region $3$\end{tabular}}}}%
    \put(0.75893519,0.35686741){\color[rgb]{0.39215686,0.39215686,0.39215686}\makebox(0,0)[lt]{\lineheight{1.25}\smash{\begin{tabular}[t]{l}Region $4$\end{tabular}}}}%
    \put(0.97062497,-0.00033175){\color[rgb]{0,0,0}\makebox(0,0)[lt]{\lineheight{1.25}\smash{\begin{tabular}[t]{l}$\orientation$\end{tabular}}}}%
    \put(-0.01617894,0.36649541){\color[rgb]{0,0,0}\makebox(0,0)[lt]{\lineheight{1.25}\smash{\begin{tabular}[t]{l}$f(\orientation)$\end{tabular}}}}%
    \put(0,0){\includegraphics[width=\unitlength,page=2]{upper_and_lower_bounding.pdf}}%
  \end{picture}%
\endgroup%

	\caption{Exemplary non-linear function and the respective piecewise linear {\color{blue}upper} and {\color{red}lower} bounds.}
	\label{fig:upper_and_lower_bounding}
	\vspace{-0.15cm}
\end{figure}

If the orientation is known, a common strategy to approximate the vehicle shape is by using three circles with radius $r$ for the rear axle, middle position, and front axle similar to \cite{Ziegler2014}, as this allows for efficient collision checking to arbitrary polygons.
We aim for a linear vehicle model, so the true (highly non-linear) orientation is unknown. As we only have an approximated orientation but do not want to underestimate any collisions, we choose to compute the upper and lower bound of the sine and cosine of the orientation. \refFigure{fig:upper_and_lower_bounding} illustrates this idea. With that and the vehicle's wheelbase $l$, we then calculate upper and lower bounds for the $x$ and $y$ position of the front axle.

\begin{subequations}
	\begin{align}
	\xfrontub &:= \x + l ~ \overline{\cos(\orientation)}\\
	\xfrontlb &:= \x + l ~ \underline{\cos(\orientation)}\\
	\yfrontub &:= \y + l ~ \overline{\sin(\orientation)}\\
	\yfrontlb &:= \y + l ~ \underline{\sin(\orientation)}
	\end{align}
	\label{eq:front_ub_lb_approx}
\end{subequations}

\begin{figure}[tb]
	\footnotesize
	\centering
	\def\svgwidth{0.65\columnwidth}
	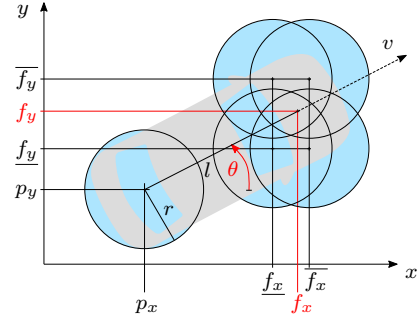
	\caption{Vehicle model with wheelbase $l$ and {\color{cyan} disk-based collision-shape} of radius $r$. The variables in red are {\color{red}unavailable} in the MIQP model formulation. The orientation $\orientation$ is defined clockwise.}
	\label{fig:collision_shape}
\end{figure}

Permuting $\xfrontub, \xfrontlb$ with $\yfrontub, \yfrontlb$ yields four circles for the front axle, which represent an over-approximation of the true front axle circle, as shown in \refFigure{fig:collision_shape}. For now, we chose to not model the middle point of the vehicle, as this increases the complexity of the model. However, a similar approach can be applied to the mid axle.
We now present two methods for obtaining bounds for the sine and cosine.

\subsubsection{Constant Approximation}
\label{subsubsec:constant_approximation}
We propose a constant approximation of the sine using the maximum and minimum orientation for each region.
\begin{subequations}
	\begin{align}
	\overline{\sin(\orientation)} &\tilde{=} \max [ \sin(\atan(\fractionparametersfour^r, \fractionparametersthree^r)), \sin(\atan(\fractionparameterstwo^r, \fractionparametersone^r)) ] \\
	\underline{\sin(\orientation)} &\tilde{=} \min [ \sin(\atan(\fractionparametersfour^r, \fractionparametersthree^r)), \sin(\atan(\fractionparameterstwo^r, \fractionparametersone^r)) ].
	\end{align}
	\label{eq:fitting_constant}
\end{subequations}
The cosine is calculated accordingly. With a higher number of regions, the error for this type of approximation will decrease. Note that this is only valid as long as a region is not defined over multiple quadrants, since sine and cosine are only monotonic functions within a quadrant.

\subsubsection{Velocity-Dependent Approximation}
\label{subsubsec:veldep_approximation}
To preserve the linearity of the constraints, only a linear combination of the state variables can be computed.
We chose to upper bound the $\sin(\orientation)$ term by a first order polynomial depending on region $r$ with three parameters $p_\square$
\begin{align}
\overline{\sin(\orientation)} \tilde{=} p_{00} + p_{10} \vx + p_{01} \vy := \polysinub(\vx, \vy)
\label{eq:fitting_polynomial}
\end{align}
and analogously fit such linear polynomials for the lower bound of the sinus $\polysinlb$ function and the bounds of the cosine $\polycoslb$ and $\polycosub$.
The methodology how we compute these parameters $p_\square$ is introduced in \refSection{sec:front_alxe_fitting}.
This approximation will lead to a front axle position that depends on the respective velocity terms. However, that is not the case if the orientation can be calculated analytically and leads to high errors for low velocities. 
\begin{figure}[t]
	\vspace{0.15cm}
	\centering
	\resizebox{.59\linewidth}{!}{\input{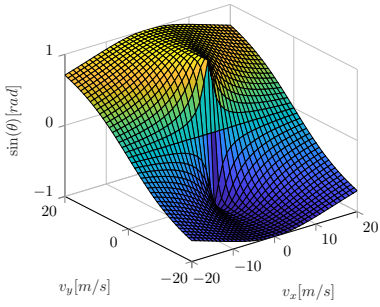}\unskip}
	\caption{Plot of $\sin(\orientation) = \sin(\atan(\vy/\vx))$ with respect to $\vx$ and $\vy$. The function is obviously highly non-linear but can be approximated in a piecewise linear fashion.}
	\label{fig:sintheta_UB}
	\vspace{-0.15cm}
\end{figure}

\subsection{Modelling the Non-Holonomics}
\label{subsec:modelling_non_holonomics}

Previous MIQP formulations \cite{Qian2016, Eilbrecht2017, Burger2018} have approximated the non-holonomics by bounding acceleration in $x$ and $y$ direction and by coupling the velocities via $\vy \in [\vx\tan(\theta_{min}), \vx\tan(\theta_{max})]$, with $\theta_{min}$ and $\theta_{max}$ being the valid orientation range of that model. However, decoupled acceleration bounds cannot yield a non-holonomic behavior.
\citet{Ziegler2014} calculate the curvature using 
\begin{equation}
\kappa = \frac{\vx\ay - \vy\ax}{\sqrt[3]{\vx^2+\vy^2}}
\label{eq:kappa}
\end{equation}
and formulate the bound constraints $\kappa \in [\kappa_{min}, \kappa_{max}]$. However, \refEquation{eq:kappa} is highly non-linear and thus curvature constraints cannot be expressed as a linear constraint for MIQP.
To obtain constraints dependent on $\ax, \ay$, we transform $\kappa_{max}, \kappa_{min}$ using \refEquation{eq:kappa} to
\begin{subequations}
\begin{align}
	\frac{\kappa_{max}}{\vx} \sqrt[3]{\vx^2+\vy^2} + \frac{\vy}{\vx}\ax \eqslantgtr \ay \label{eq:kappa_max} \\
	\frac{\kappa_{min}}{\vx} \sqrt[3]{\vx^2+\vy^2} + \frac{\vy}{\vx}\ax \eqslantless \ay. \label{eq:kappa_min}
\end{align}
\label{eq:kappa_holonomy}
\end{subequations}
We can then apply the concept of region-wise linearization described in \ref{subsec:disjunctive_modelling} to obtain linear constraints and fit two linear polynomials for upper $\polykappaub$ and lower $\polykappalb$ bounding the curvature as shown in \refSection{sec:curvature_fitting}.

\section{Fitting Method}
\label{sec:method_fitting}
In this section, we will briefly describe how we fit the parameters in \ref{subsubsec:veldep_approximation} and \ref{subsec:modelling_non_holonomics}.
All fits are done on a region-wise basis and yield polynomials of the form of \refEquation{eq:fitting_polynomial}.

\subsection{Fitting the Front axle Position}
\label{sec:front_alxe_fitting}
\begin{figure}[t]
	\vspace{0.15cm}
	\centering
	\resizebox{.8\linewidth}{!}{\input{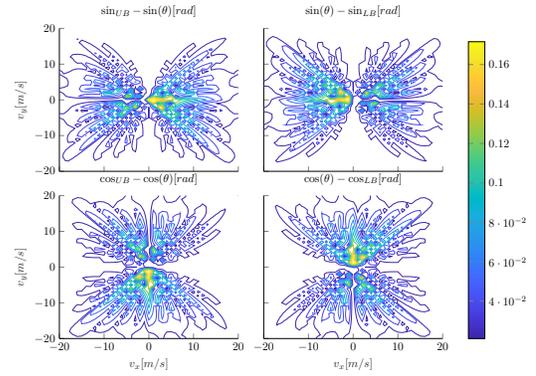}\unskip}
	\caption{Errors of the piecewise linear fitting using 32 regions of upper $\square_{UB}$ and lower bounds $\square_{LB}$ trigonometric functions.}
	\label{fig:error_trigonometrics}
	\vspace{-0.15cm}
\end{figure}

In \refSection{subsec:overapproximation_collision_shape}, we motivated the need to linearize the trigonometric functions \refEquation{eq:trigonometrics_sin} and \refEquation{eq:trigonometrics_cos}, which depend on $\vx$ and $\vy$. 
Both sine and cosine are highly non-linear, in \refFigure{fig:sintheta_UB} we show the sine function.
We formulate the problem to find a piecewise upper bound to a two-dimensional nonlinear function of $\vx$ and $\vy$ as a linear least-squares problem with linear constraints.
\refFigure{fig:error_trigonometrics} shows the errors we obtain from the fitting. 
For the upper and lower bounds of the sine and cosine function the orientation error is always below $0.16$ rad.
Due to 
\begin{subequations}
\begin{align}
sin(\vx = 0, \vy \to \pm 0) &= \pm \infty \\
cos(\vx \to \pm 0, \vy = 0) &= \pm \infty,
\end{align}
\label{eq:sin_cos_limits}
\end{subequations}
we get the highest error close to the origin.
For higher velocities, the errors are significantly smaller due to a better approximation of the non-linear function.

\refTable{tab:table_orientation_evaluation} shows the positional error of the front axle for a range of orientations in the first quadrant in comparison to the constant approximation (denoted by const.).
We observe that the upper and lower bound always have different signs, which means that the actual axle position is always within that. This shows the validity of the over-approximation of the front axle. 
As expected, the error becomes smaller with an increasing number of regions, as we are using smaller linear pieces to approximate the non-linear functions. 
In general, the error of the velocity-dependent approximation is smaller than for the constant approximation. 
For velocities $\lessapprox 0.1 m/s$, a constant approximation yields smaller errors, as motivated by \refEquation{eq:sin_cos_limits}.
Implementing an optimal transition strategy between the two approximations is the subject of future work.
\begin{table}[t]
	\vspace{0.15cm}
	\centering
	\caption{Absolute positional errors $(\xfront - [\xfrontlb, \xfrontub])$ and $(\yfront - [\yfrontlb, \yfrontub])$ for the approximation of the front rear axle in meters.}
	\scriptsize
	\begin{tabular}{l|l|l|l|l|l|l}
$\orientation$ & $v$ &  & 16 regions & 32 regions & 64 regions & 128 regions\\
\hline
\multirow{6}{*}{$ 0\si{\degree}$}& \multirow{2}{*}{const.}& x & 0.21, 0.00 & 0.05, 0.00 & 0.01, 0.00 & 0.00, 0.00\\
& & y & 0.00, -1.07 & 0.00, -0.55 & 0.00, -0.27 & 0.00, -0.14\\
\cline{2-7}
& \multirow{2}{*}{$ 0.1\frac{m}{s} $} & x & 0.20, -0.01 & 0.07, -0.01 & 0.01, -0.01 & 0.01, -0.01\\
& & y & 0.01, -0.97 & 0.01, -0.52 & 0.01, -0.09 & 0.01, -0.01\\
\cline{2-7}
& \multirow{2}{*}{$ 20\frac{m}{s} $} & x & 0.01, -0.08 & 0.01, -0.03 & 0.01, -0.01 & 0.01, -0.01\\
& & y & 0.01, -0.01 & 0.01, -0.01 & 0.01, -0.01 & 0.01, -0.01\\
\hline
\multirow{6}{*}{$ 45\si{\degree}$}& \multirow{2}{*}{const.}& x & 0.00, -0.61 & 0.00, -0.35 & 0.00, -0.18 & 0.00, -0.09\\
& & y & 0.91, 0.00 & 0.42, 0.00 & 0.20, 0.00 & 0.10, 0.00\\
\cline{2-7}
& \multirow{2}{*}{$ 0.1\frac{m}{s} $} & x & 0.01, -0.61 & 0.01, -0.36 & 0.01, -0.17 & 0.01, -0.03\\
& & y & 0.89, -0.01 & 0.43, -0.01 & 0.18, -0.01 & 0.03, -0.01\\
\cline{2-7}
& \multirow{2}{*}{$ 20\frac{m}{s} $} & x & 0.04, -0.19 & 0.02, -0.12 & 0.02, -0.06 & 0.01, -0.02\\
& & y & 0.27, -0.11 & 0.14, -0.05 & 0.06, -0.02 & 0.02, -0.01\\
\hline
\multirow{6}{*}{$ 90\si{\degree}$}& \multirow{2}{*}{const.}& x & 0.00, -1.07 & 0.00, -0.55 & 0.00, -0.27 & 0.00, -0.14\\
& & y & 0.21, 0.00 & 0.05, 0.00 & 0.01, 0.00 & 0.00, 0.00\\
\cline{2-7}
& \multirow{2}{*}{$ 0.1\frac{m}{s} $} & x & 0.01, -0.98 & 0.01, -0.53 & 0.01, -0.10 & 0.01, -0.01\\
& & y & 0.20, -0.01 & 0.07, -0.01 & 0.01, -0.01 & 0.01, -0.01\\
\cline{2-7}
& \multirow{2}{*}{$ 20\frac{m}{s} $} & x & 0.01, -0.01 & 0.01, -0.01 & 0.01, -0.01 & 0.01, -0.01\\
& & y & 0.01, -0.08 & 0.01, -0.03 & 0.01, -0.01 & 0.01, -0.01\\
\end{tabular}

	\label{tab:table_orientation_evaluation}
	\vspace{-0.15cm}
\end{table}

\subsection{Fitting the Curvature}
\label{sec:curvature_fitting}
As we discussed in \refSection{subsec:modelling_non_holonomics}, we approximate the curvature constraints by \refEquation{eq:kappa_holonomy}. We choose to fit the polynomials $\polykappa$ on 
\begin{subequations}
	\begin{align}
	\frac{\kappa_{max}}{\vx} \sqrt[3]{\vx^2+\vy^2} &\equiv \polykappaub \geq \ay - \frac{\vy}{\vx}\ax\label{eq:kappa_max_prime} \\
	\frac{\kappa_{min}}{\vx} \sqrt[3]{\vx^2+\vy^2} &\equiv \polykappalb \leq \ay - \frac{\vy}{\vx}\ax \label{eq:kappa_min_prime}
	\end{align}
	\label{eq:kappa_min_prime_both}
\end{subequations}
and use these polynomials in inequality constraints bounding $\axy$ as \refEquation{eq:kappa_min_prime_both} indicates.
We approximate the term $\frac{\vy}{\vx}$ in the MIQP formulation by the mean orientation within the respective region, using the region boundaries, see \refEquation{eq:region_linear_equations}. 
The larger the regions (the fewer number of regions), the higher the error will be.
We then solve two unconstrained linear least-square problems by minimizing the error to \refEquation{eq:kappa_max_prime}, yielding the linear polynomial $\polykappaub$, and \refEquation{eq:kappa_min_prime}, yielding $\polykappalb$.

\section{Formulating the Planning Problem as Mixed-Integer Quadratic Problem}
\label{sec:method_miqp}
We model the planning problem as MIQP without restricting the validity scope of the model.
The vehicle motion model is formulated as discrete linear constraints.
Using binary variables, we ensure collision-freeness and a correct non-holonomic motion of the vehicle.
A quadratic objective function keeps the solution close to the reference.

Subsequently, we use the nomenclature for the decision variables introduced in \refTable{tab:nomenclature_variables}.
The subscript $\square_{ref}$ denotes the respective reference.
We optimize a discrete time range from $\t_2$ to $\t_N$ with $\deltat$ increment.
By $\timeintervall$, we denote the time interval $[\t_1, \dots, \t_N]$.
All decision variables are initialized with the current state of the vehicle at $\t_1$.
We bound the speed by the minimal and maximal values $\vmin$ and $\vmax$ from the fitting, as our approximations are only valid there.
The subscript $\xy$ denotes the respective term for both, $x$ and $y$ direction.
In the following, we will use this notation for compactness of the equations wherever it does not lead to ambiguousness.

\begin{table}[t]
\vspace{0.15cm}
\caption{Decision variables used throughout this paper, with discrete time dependency $\t$, region dependency $r$, environment dependency $\envpolyline$, and obstacle dependency $\obstacle$.}
\centering
\begin{tabularx}{8cm}{r|X|l} %
	Variable & Description & Range\\ 
	\hline 
	$\x(\t)$, $\y(\t)$ & vehicle rear axle center position  & free\\ 
	$\vx(\t)$, $\vy(\t)$ & vehicle velocity in $x$, $y$ direction & $[\vmin,\vmax]$\\
	$\ax(\t)$, $\ay(\t)$ & vehicle acceleration in $x$, $y$ direction & $[\amin,\amax]$\\
	$\ux(\t)$, $\uy(\t)$ & vehicle jerk in $x$, $y$ direction & $[\umin,\umax]$\\
	$\xfrontub(\t)$, $\yfrontub(\t)$ & upper bound of the vehicle front axle center position & free \\
	$\xfrontlb(\t)$, $\yfrontlb(\t)$ & lower bound of the vehicle front axle center position & free \\
	$\region(\t,r)$ & region $r$ the vehicle is in at time $\t$ & binary\\
	$\noregionchange(\t)$ & no region change allowed & binary\\
	$\notwithinenv(\t,\envpolyline)$ & vehicle is not inside the environment sub-polygon $\envpolyline$ at time $t$ & binary\\
	$\deltacc(\t,\obstacle)$ & vehicle does not collide with obstacle $\obstacle$ at time $t$ at the reference point $p$ & binary\\
\end{tabularx} 
\label{tab:nomenclature_variables}
\vspace{-0.15cm}
\end{table}

\subsection{Formulating Reference Tracking as Objective Function}
As cost function $J$, we chose the weighted sum of position and velocity distance to the reference. 
For acceleration and jerk, we aim to minimize the squared values to avoid changes.
Suitable cost terms $q_\square$ balance the solution.
\begin{align}
J = \sum_{\t \in \timeintervall} \big ( &q_{p} (\x(\t) - \xref(\t))^2 + q_{p} (\y(\t) - \yref(\t))^2 \nonumber \\
  + ~& q_{v} (\vx(\t) - \vxref(\t))^2 + q_{v} (\vy(\t) - \vyref(\t))^2 \nonumber \\
  + ~& q_{a} \ax(\t)^2 + q_{a} \ay(\t)^2 + q_{u} \ux(\t)^2 + q_{u} \uy(\t)^2 \big )
  \label{eq:objective}
\end{align}
To formulate a MIQP problem, the objective function has to be a sum of squared or linear terms.
Therefore, we cannot use the absolute velocity (or acceleration) in the objective, since with $|v| = \sqrt{(\vx^2+\vy^2)}$, the term $(|v|^2 - |v_{ref}|^2)^2$ is not quadratic.
Similarly, costs on the angular velocity would result in non-quadratic terms. %
Furthermore, as we do not calculate a distance to objects in the model, we cannot use these distance terms in the cost function. 
Note that with this formulation of the objective function we track a reference trajectory and not a reference path.

\subsection{Formulating the Vehicle Model as Constraints}
\label{sec:vehicle_model_constraints}
The vehicle dynamics are defined by:
\begin{align}
&\begin{bmatrix}
\pxy(\t_{i+1})\\
\vxy(\t_{i+1})\\
\axy(\t_{i+1})
\end{bmatrix}
\!\!= \!\!\begin{bmatrix}
1 & \!\!\!\deltat &\!\!\!\deltat^2/2 \\
0 & \!\!\!1 & \!\!\!\deltat \\
0 & \!\!\!0 & \!1
\end{bmatrix}
\!\!\begin{bmatrix}
\pxy(\t_{i})\\
\vxy(\t_{i})\\
\axy(\t_{i})
\end{bmatrix}
\!\!+ \!\! 
\begin{bmatrix}
\deltat^3/6\\
\deltat^2/2\\
\deltat
\end{bmatrix}
\!\uxy(\t_{i}) \nonumber \\
&\forall \t_i \in [\t_1,\dots,\t_{N-1}] \label{eq:vehicle_model}
\end{align}
With this linear model correct non-holonomic motion as well as correct acceleration and steering angle limits are only valid around a small reference orientation. 
We overcome this property by introducing validity \textit{regions}, see \refSection{subsec:disjunctive_modelling}.
The set of regions covering the full orientation of $360$ degrees is denoted by $\setofregions$.
By the superscript $\square^r$, we denote region-dependent parameters in the following.
All region-dependent parameters are summarized in \refTable{tab:nomenclature_fittingparameters}.

We set the binary decision variable $\region(\t,r)$ defining in which region $r$ the vehicle is in at time $\t$ according to the two lines defining the region, cf. \refEquation{eq:region_linear_equations}.
We force the solution to lie in exactly one region.
\begin{align}
\sum_{r \in \setofregions} \region(\t,r) = 1 ~  \forall \t \in \timeintervall \label{eq:only_one_region}
\end{align}

Since with an increasing number of regions, the evaluation time of the model increases as well, we a-priori restrict the optimization to only use a set of \textit{allowed} regions and pre-compute a parameter $\possibleregion$ accordingly.
Non-allowed regions are those that cannot physically be reached anyway within the planning horizon.
To implement logical constraints, we use the well-know technique of introducing a big constant $M$ to switch inequalities.
The intuitive explanation of the now often used term $M (1-\region(\t,r))$ is "this equation is active if and only if region $r$ is active at time-step $\t$".
The active region is set by the following set of constraints:

\begin{subequations}
\begin{align}
\fractionparametersone^r \vy(\t) \geq \fractionparameterstwo^r \vx(\t) - M (1-\region(\t,r)) \label{eq:select_region_low}\\
\fractionparametersthree^r \vy(\t) \leq \fractionparametersfour^r \vx(\t) + M (1-\region(\t,r)) \label{eq:select_region_up}\\
 \forall \t \in \timeintervall, ~ \forall r \in \setofregions, ~ \text{if} ~ \possibleregion(r) = 1 \nonumber
\end{align}
\label{eq:speed_select_active_region}
\end{subequations}

Regions marked as non-possible by $\possibleregion$ may not be selected.
With this implementation, the model formulation is generic for all scenarios.
\begin{align}
\region(\t,r) = 0 ~  \forall \t \in \timeintervall, ~ \forall r \in \setofregions, ~ \text{if} ~ \possibleregion = 0 \label{eq:speed_select_active_region_not_allowed}
\end{align}

We impose region-dependent limits on acceleration and jerk to make sure we always meet the correct absolute possible acceleration and jerk.
As acceleration $\axy$ and jerk $\uxy$ are defined in a global system, the limits are rotated with the regions.
In \refEquation{eq:state_limits}, we only state the constraints for the acceleration, the formulation for jerk is alike.
Note that the speeds are naturally bounded correctly by \refEquation{eq:speed_select_active_region}.
\begin{subequations}
\begin{align}
\axy(\t) &\leq \amaxxy + M (1-\region(\t,r))\\
\axy(\t) &\geq \aminxy - M (1-\region(\t,r))\\
 \forall \t \in \timeintervall, ~ &\forall r \in \setofregions, ~ \text{if} ~ \possibleregion = 1 \nonumber  
\end{align}
\label{eq:state_limits}
\end{subequations}

\subsection{Modeling the Non-Holonomy as Constraints}
To ensure a correct non-holonomic movement of the vehicle, we limit the lateral acceleration using the maximal available curvature per region as motivated in \refSection{subsec:modelling_non_holonomics}.
We fit lower and upper linear approximation polynomials  $\polykappaub$ and $\polykappalb$ dependent on $\vx$, $\vy$, and the region $r$.
The following constraints model the inequalities bounding the curvature $\kappa$ as stated in \refEquation{eq:kappa_holonomy}.
\begin{subequations}
\begin{align}
\ay(\t) - \frac{\fractionparameterstwo^r+\fractionparametersfour^r}{\fractionparametersone^r+\fractionparametersthree^r} \ax(\t) \leq\nonumber\\ \polykappaub(\vx(\t), \vy(\t)) + M (1-\region(\t,r)) + M \noregionchange(\t) \\
\ay(\t) - \frac{\fractionparameterstwo^r+\fractionparametersfour^r}{\fractionparametersone^r+\fractionparametersthree^r} \ax(\t) \geq\nonumber\\ \polykappalb(\vx(\t), \vy(\t)) + M (1-\region(\t,r)) + M \noregionchange(\t) \\
 \forall \t \in \timeintervall, ~ \forall r \in \setofregions, ~ \text{if} ~ \possibleregion = 1 \nonumber  
\end{align}
\label{eq:curvature_constraint}
\end{subequations}

At very low vehicle speeds, too tight curvature constraints limit the acceleration so that other accelerations than zero are not possible. In contrast, too loose curvature constraints will violate the non-holonomy.
We therefore introduce a minimum speed limit.
If either $|\vx|$ or $|\vy|$ is below that limit, regions changes are not allowed, which we indicate by setting the binary helper variable $\noregionchange$ to true.
\begin{subequations}
\begin{align}
\region(\t_i,r) - \region(\t_{i-1},r) \leq 1-\noregionchange(\t_i)\\
\region(\t_i,r) - \region(\t_{i-1},r) \geq -1+\noregionchange(\t_i)\\
\forall \t_i \in [\t_2,\dots,\t_N], ~ \forall r \in \setofregions \nonumber 
\end{align}
\label{eq:low_speed_constraint}
\end{subequations}
\subsection{Approximating the Front Axle Position as Constraints}
In \refSection{subsec:overapproximation_collision_shape} we introduced the concept of how to approximate lower and upper bounds for the position of the front axle of the vehicle.
Using this idea, we formulate constraints performing an (over-)approximative collision check for the front axle instead of computing the intersection or distance of the actual vehicle shape with obstacles or the environment.
We calculate the lower and upper bounds $\squarefrontlb$, $\squarefrontub$ of the true, unknown, front position $(\xfront, \yfront)$ with respect to the current region $r$ where $l$ denotes the wheelbase of the vehicle.
We only state the equations for the upper bounds here, the lower bound constraints are formulated alike.
\begin{subequations}
	\begin{align}
	\!M\!(\region(\t,r){-}1) \!\leq\! \xfrontub(\t)\!-\!\x(\t)\!-\!l\polycosub(\vx(\t),\vy(\t))\\
	\!M\!(1{-}\region(\t,r)) \!\geq\! \xfrontub(\t)\!-\!\x(\t)\!-\!l\polycosub(\vx(\t),\vy(\t))\\
	\!M\!(\region(\t,r){-}1) \!\leq\! \yfrontub(\t)\!-\!\y(\t)-l\polysinub(\vx(\t),\vy(\t))\\
	\!M\!(1{-}\region(\t,r)) \!\geq\! \yfrontub(\t)\!-\!\y(\t)\!-\!l\polysinub(\vx(\t),\vy(\t))\\
	\forall \t_i \in \timeintervall, ~ \forall r \in \setofregions, ~ \text{if} ~ \possibleregion(r) = 1 \nonumber  
	\end{align}
	\label{eq:front_ub_lb}
\end{subequations}
\subsection{Constraints Limiting the Model to Stay on the Road}
\label{sec:constaints_environment}

\begin{figure}[tb]
	\vspace{0.15cm}
	\footnotesize
	\centering
	\def\svgwidth{0.58\columnwidth}
	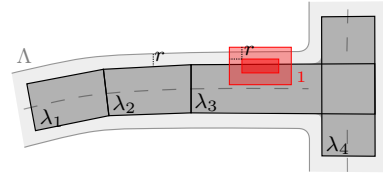
	\caption{Schematic sketch showing  how the environment polygon $\envpolys$ is shrieked by the collision circle radius $r$ and split into several convex polygons $\envpolyline_\square$. Obstacles $\obstacle_\square$ are also inflated with $r$.}
	\label{fig:environment_polygon_inflation_and_convex}
	\vspace{-0.15cm}
\end{figure}

This section introduces how we enforce the vehicle to stay within an environment modeled as an arbitrary, potentially non-convex closed polygon $\envpolys$ \cite{Frese2011}.  
The environment is deflated with the radius of the collision circles, see \refFigure{fig:environment_polygon_inflation_and_convex}.
Non-convex environment polygons are split into several convex sub-polygons $\envpolyline$. 
We enforce the vehicle to be in at least one of these convex sub-polygons.
These sub-polygons are represented by a set of line segments, denoted by $\envpolylinesegment$, between two points $\envpolylinesegmentpointone^\envpolylinesegment$ and $\envpolylinesegmentpointtwo^\envpolylinesegment$.
With this strategy, the polygon-to-polygon collision check narrows down to a point-to-polygon check.

We enforce the rear axle position $\pxy$ and the lower and upper bounds of the front axle position, namely the four points $(\xfrontlb, \yfrontlb)$, $(\xfrontlb, \yfrontub)$, $(\xfrontub, \yfrontlb)$, and $(\xfrontub, \yfrontub)$, to be within the environment polygon.
Each set of constraints is formulated in a similar manner, \refEquation{eq:collision_environment} states the equations for the point $(\x, \y)$.
The decision variable $\notwithinenv$ models that all five points do not collide with the environment sub-polygon $\envpolyline$ at time $\t$.
By $\envpolylinesegment_\medstar$, we here denote $\envpolylinesegmentpointtwo_\medstar^\envpolylinesegment - \envpolylinesegmentpointone_\medstar^\envpolylinesegment$ for $x$ and $y$ respectively.
\begin{align}
\envpolylinesegment_x (\y(\t) - \envpolylinesegmentpointone_y^\envpolylinesegment) - 
\envpolylinesegment_y (\x(\t) - \envpolylinesegmentpointone_x^\envpolylinesegment) &\leq -M \notwithinenv(\t, \envpolyline) \label{eq:collision_environment}\\
 \forall \t \in \timeintervall, ~ \forall \envpolylinesegment \in \envpolyline, ~ \forall \envpolyline \in \envpolys \nonumber 
\end{align}

To ensure that all vehicle points are at least within one of the environment polygons, we set
\begin{align}
\sum_{\envpolyline \in \envpolys }\notwithinenv(\t, \envpolyline) \leq |\envpolys| - 1 ~  \forall \t \in \timeintervall,
\label{eq:envpoly_sum}
\end{align}
where $|\envpolys|$ denotes the number of convex sub-environments.

\subsection{Formulating Collision Avoidance as Constraints}
We enforce the vehicle to not collide with an arbitrary number of static or dynamic convex obstacle polygons $\obstacleset$.
Non-convex obstacle shapes as described in \refSection{sec:constaints_environment} split into several convex ones.
The obstacles are inflated with the radius of the collision circles, cf. \refFigure{fig:environment_polygon_inflation_and_convex}.

$\envpolylinesegment$ again denotes one line segment of one obstacle $\obstacle$ within the set of obstacles $\obstacleset$ with startpoint  $\envpolylinesegmentpointone^l$ and endpoint $\envpolylinesegmentpointtwo^l$. 
We enforce the vehicle rear axle position $\pxy$ and the four permutations of the front axle bounds to be collision-free.
In contrast to the environment, which we assume as constant over time, the obstacle polygons may vary their position and shape over time, but preserve the polygon topology.
By decision variables $\deltaccall$, we indicate whether none of the five points collide with the obstacle $\obstacle$.
\refEquation{eq:collions_obstacle} states the inequalities for the point $(\x,\y)$ constraining $\deltacc$. 
The four points representing the collision shape approximation of the front axle $\xyfront$ are each taken into account by four more sets of similar decision variables $\deltaccall$ and sets of inequalities. 

\begin{align}
\envpolylinesegment_x (\y(\t) - \envpolylinesegmentpointone_y^\envpolylinesegment) - 
\envpolylinesegment_y (\x(\t) - \envpolylinesegmentpointone_x^\envpolylinesegment) &\leq M \deltacc(\t, l) \label{eq:collions_obstacle}\\
 \forall \t \in \timeintervall, ~ \forall \envpolylinesegment \in \obstacle, ~ \forall \obstacle \in \obstacleset \nonumber 
\end{align}

Denoting the number of line segments in a sub-polygon by $|\obstacle|$, we enforce each of the five points to not lie within an obstacle.
\begin{align}
\sum_{\envpolylinesegment \in \obstacle} \deltacc(\t, \envpolylinesegment) \leq |\obstacle|-1 \label{eq:collions_obstacle_sums} ~ %
 \forall \t \in \timeintervall, ~ \forall \obstacle \in \obstacleset %
\end{align}
\subsection{Optimization Problem}
Collecting all constraints from above, the final optimization problem can be written as
\begin{align}
\text{minimize } &(\ref{eq:objective}) \nonumber\\
\text{subject to } &(\ref{eq:vehicle_model}),
 (\ref{eq:only_one_region}),
 (\ref{eq:speed_select_active_region}),
 (\ref{eq:speed_select_active_region_not_allowed}),
 (\ref{eq:state_limits}),
 (\ref{eq:curvature_constraint}), \nonumber\\
 &(\ref{eq:low_speed_constraint}),
 (\ref{eq:front_ub_lb}), 
 (\ref{eq:collision_environment}), 
 (\ref{eq:envpoly_sum}), 
 (\ref{eq:collions_obstacle}), 
 (\ref{eq:collions_obstacle_sums}) \nonumber
\end{align}
The formulation is a standard MIQP model that can be solved with an off-the-shelf solver.
For a faster optimization solution, we bound acceleration $\axy$ and jerk $\uxy$ with constant values $\amin$, $\amax$ and $\umin$, $\umax$ respectively. We chose these values as minima and maxima of the region-dependent limits $\aminxy$, $\amaxxy$ and $\uminxy$, $\umaxxy$ respectively.

\section{Evaluation}
\label{sec:evaluation}
We will now evaluate our model in two different scenarios. 
We prove in \refSection{seq:eval:nonholonomy}, that the result of the optimization is a drivable trajectory for a non-holonomic vehicle model.
\refSection{sec:eval:obstacles} show the performance in an obstacle avoidance scenario. 
We only apply cost term on reference position tracking. We use a model with 32 regions and velocity-dependent front axle approximation. 

\subsection{Preserving the Non-Holonomy}
\label{seq:eval:nonholonomy}
To show the effectiveness of our constraints guaranteeing non-holonomic motions, we optimize two different reference trajectories, both forming a circle to perform a 90-degree turn. 
While it is preferable to only generate reference trajectories with only valid curvatures, this example clearly shows that our model preserves the non-holonomy. 
The same property is also necessary for obstacle avoidance.
As baseline algorithms for comparison we implemented two SQP-based optimization models, one using a standard bicycle model and another using the same triple integrator as our MIQP model but with a nonlinear curvature constraint, following \refEquation{eq:kappa}.

\begin{figure}[t]
	\vspace{0.15cm}
	\centering
	\resizebox{.9\linewidth}{!}{\input{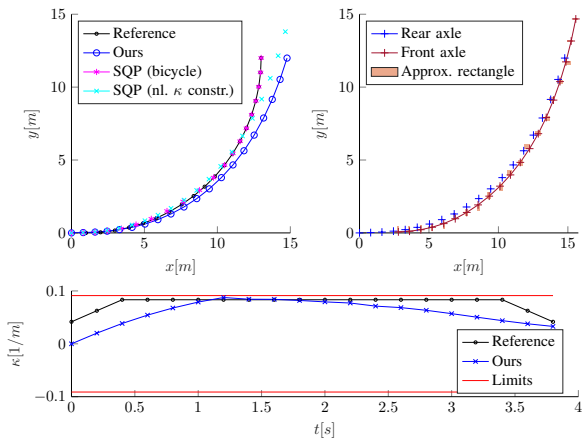}\unskip}
	\caption{Feasible reference: Our model stays within the curvature bounds (lower) and shows good trajectory tracking behavior as the reference implementations using an SQP optimizer does (upper left). 
		The true front axle position is always within the lower/upper bound approximation rectangles of the front axle (upper right).}
	\label{fig:circle_feasible}
	\vspace{-0.15cm}
\end{figure}

\begin{figure}[t]
	\vspace{0.15cm}
	\centering
	\resizebox{.9\linewidth}{!}{\input{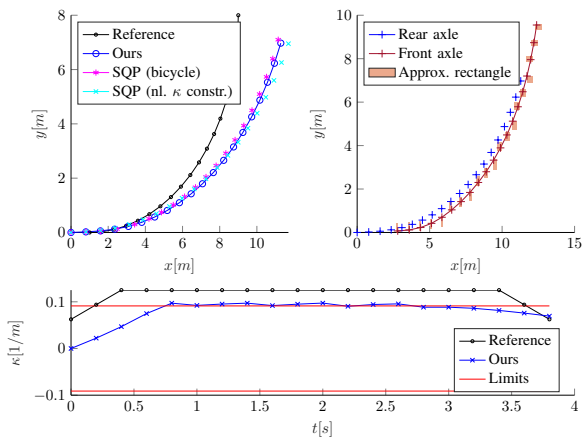}\unskip}
	\caption{Infeasible reference: All three optimization solutions cannot track the reference despite operating at maximum curvature.}
	\label{fig:circle_infeasible}
	\vspace{-0.15cm}
\end{figure}

\refFigure{fig:circle_feasible} shows the results for a turning radius that our vehicle model is able to follow. Our optimization yields a trajectory that closely follows the reference.
However, the second reference in \refFigure{fig:circle_infeasible} models a turning radius that is too small (the curvature of the reference exceeding the limits). 
As desired, our model does not follow the reference and yields a trajectory that stays within the curvature bounds. 
As we fit the curvature approximation polygons in the mean of each region (see  \refSection{sec:curvature_fitting}), we can slightly exceed the curvature bound at region boundaries. This can easily be mitigated using a safety margin.
\subsection{Avoiding Dynamic Obstacles within the Road Boundaries}
\label{sec:eval:obstacles}
\begin{figure}[t]
	\centering
	\vspace{0.15cm}
	\resizebox{\linewidth}{!}{\input{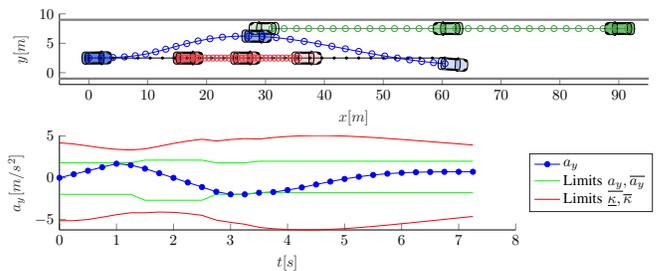}\unskip}
	\caption{An Overtaking scenario with oncoming traffic. The upper figure shows the trajectories for the ego vehicle (blue) and the other traffic participants (red, green). $a_y$ of the planned trajectory stays within the region-dependent acceleration bounds as well as the bounds approximating the curvature limit.}
	\vspace{-0.15cm}
	\label{fig:dynamic_obstacles}
\end{figure}

The second example considers a two-lane road, where the ego vehicle drives at the speed of $30km/h$ and approaches a slower vehicle traveling at $10km/h$. There is oncoming traffic in the other lane traveling at $30 km/h$. 
The reference trajectory represents the center line of the right lane traveling at the reference speed. 
We obtain deterministic predictions of those two traffic participants and include them as dynamic obstacles in our optimization. 
\refFigure{fig:dynamic_obstacles} shows that the optimizer is able to find a trajectory that overtakes the red vehicle in front and goes back to the right lane to avoid the green, oncoming vehicle. 
We observe that the acceleration in the $y$-direction stays within the acceleration limits. 
It is much closer than the curvature-induced acceleration limit following \refEquation{eq:curvature_constraint}, which is reasonable, as lateral movement in straight scenarios traveling at moderate or high velocities is constrained by the inertia of the vehicle, not the non-holonomics.

\section{Conclusion and Future Work}
\label{sec:conclusion}
We introduced a novel mixed-integer formulation for the behavior planning problem.
We showed the correct non-holonomic motion of our optimized result for arbitrary road curvatures and arbitrary orientations of the vehicle.
Using linear overapproximations of the collision shape, we enable the MIQP formulation to not cause any collisions.
This provides us with a solution method, that obtains kinematically valid, globally optimal solutions avoiding obstacles without any form of randomness. 
This property is certainly a favorable feature for validation and certification.
This will enable us in the future to apply the MIQP behavior planning model in real-road driving scenarios using our institute's research vehicle \cite{Kessler2019a}.
Future work will expand this framework to the multi-agent case for cooperative planning as well as to investigate the benefits from logical constraints for the formulation of traffic rules.

\section*{Acknowledgment}
This research was funded by the Bavarian Ministry of Economic Affairs, Regional Development and Energy, project Dependable AI and supported by the Intel Collaborative Research Institute - Safe Automated Vehicles.

\renewcommand{\bibfont}{\footnotesize}
\printbibliography
\end{document}